\documentclass[letterpaper, 10 pt, conference]{ieeeconf}  

\IEEEoverridecommandlockouts                              
\usepackage{graphics} 
\usepackage{epsfig} 
\usepackage{mathrsfs}
\usepackage{times} 
\usepackage{amsmath} 
\usepackage{amssymb}  
\usepackage{algpseudocode}
\usepackage{algorithm}
\usepackage{dblfloatfix}    
\usepackage{optidef}

\algnewcommand{\algorithmicgoto}{\textbf{go to}}%
\algnewcommand{\Goto}[1]{\algorithmicgoto~\ref{#1}}

\title{\LARGE\bf
Energy-Efficient Motion Planning for Multi-Modal Hybrid Locomotion}
\author{H.J. Terry Suh\textsuperscript{1}, Xiaobin Xiong\textsuperscript{2}, Andrew Singletary\textsuperscript{2}, Aaron D. Ames\textsuperscript{2}, Joel W. Burdick\textsuperscript{2}
\thanks{\textsuperscript{*} This work is supported by NSF Award No. 1932091}
\thanks{\textsuperscript{1}Computer Science and Artificial Intelligence Laboratory, Massachusetts Institute of Technology, Cambridge, MA 02139, USA, {\tt\small hjsuh@mit.edu}}%
\thanks{\textsuperscript{2}Deptartment of Mechanical and Civil Engineering, California Institute of 
Technology, Pasadena, CA 91125, USA, {\tt\small\{xxiong,asinglet,}\newline {\tt\small ames\}@caltech.edu, jwb@robotics.caltech.edu}  }
}

\usepackage{graphicx}

\usepackage[strings]{underscore}
\usepackage{float}
\usepackage{setspace}
\usepackage{mdframed}
\usepackage{hyperref}
\hypersetup{
    colorlinks=true,
    linkcolor=blue,
    filecolor=green,      
    urlcolor=black,
}
\newcommand{\note}[1]{}

\begin{document}

\maketitle
\thispagestyle{empty}
\singlespace
\setcounter{tocdepth}{2}

\begin{abstract}
Hybrid locomotion, which combines multiple modalities of locomotion within a single robot, enables robots to carry out complex tasks in diverse environments. This paper presents a novel method for planning multi-modal locomotion trajectories using approximate dynamic programming.  We formulate this problem as a shortest-path search through a state-space graph, where the edge cost is assigned as optimal transport cost along each segment. This cost is approximated from batches of offline trajectory optimizations, which allows the complex effects of vehicle under-actuation and dynamic constraints to be approximately captured in a tractable way. Our method is illustrated on a hybrid double-integrator, an amphibious robot, and a flying-driving drone, showing the practicality of the approach.
\end{abstract}

\vskip -0.2 true in
\section{Introduction}\label{introduction}

A hybrid locomotor combines multiple movement modalities into a single platform. Examples of hybrid locomotion include amphibious vehicles with the ability to swim and drive, or flying cars with the ability to drive and fly. Hybrid locomotion can allow robots to tackle more complex tasks in complicated environments, while achieving greater performance, such as improved energy efficiency. For instance, a flying-car can readily fly over obstacles or uneven terrain via aerial mobility, while driving when possible to save energy (see Fig. 1 for examples).


Prior works on hybrid locomotion  have investigated the design and feasibility of hybrid locomotion strategies (\cite{Ambot,aquabot,picobug2,kevin,robotbee,flyingstar}); however, realizing the full potential of these robots not only depends on clever design, but also on autonomous planning of their complex motion strategies. The continuous inputs, combined with discrete mode switches, produce entirely different energy costs, travel times, and robustness, which ultimately dictate performance. The complexity of combinatorial optimization of the switching sequences, as well as trajectory optimization within each modality, makes this problem particularly challenging. Directly transcribing this problem into a mixed-integer program \cite{mixedinteger,tobia} may not scale well enough to handle switching sequences and coordinates of high-dimensional problems. 


This paper presents a novel motion-planning method for hybrid locomotion using approximate dynamic programming.  Our solution relies on the following key insight: jointly optimizing for the continuous and discrete decision variables in the space of policies is difficult, but we can use the approximate optimal cost of a continuous trajectory segment as a proxy to the effect of optimal continuous policies within each modality. This allows us to decouple the original mixed-integer problem into discrete and continuous problems by first performing a graph search with approximated energy costs, then doing a final multi-phase continuous optimization with the obtained sequence and switch coordinates.


Prior work on multi-modal planning often only consider geometric graph-based planning (\cite{hybridswarm},\cite{amphibiousplanning},\cite{HyFDR},\cite{Hauser2010}). By ignoring the continuous dynamics of the robot, these planners often ignore dynamic feasibility. In addition, although energy expenditure is often the most relevant cost in hybrid locomotion, most works often assume that energy is linearly proportional to geometrical distance traveled (\cite{hybridswarm},\cite{HyFDR},\cite{IVS}), which ignore how dynamic constraints affect the optimal cost. As our cost is approximated from full-dimensional trajectory optimization with rigid body and motor dynamics, it accurately captures the robot's dynamic characteristics, including effects of underactuation that might occur in some locomotion modalities.

\begin{figure}[t]
	\centering\includegraphics[width = 0.45\textwidth]{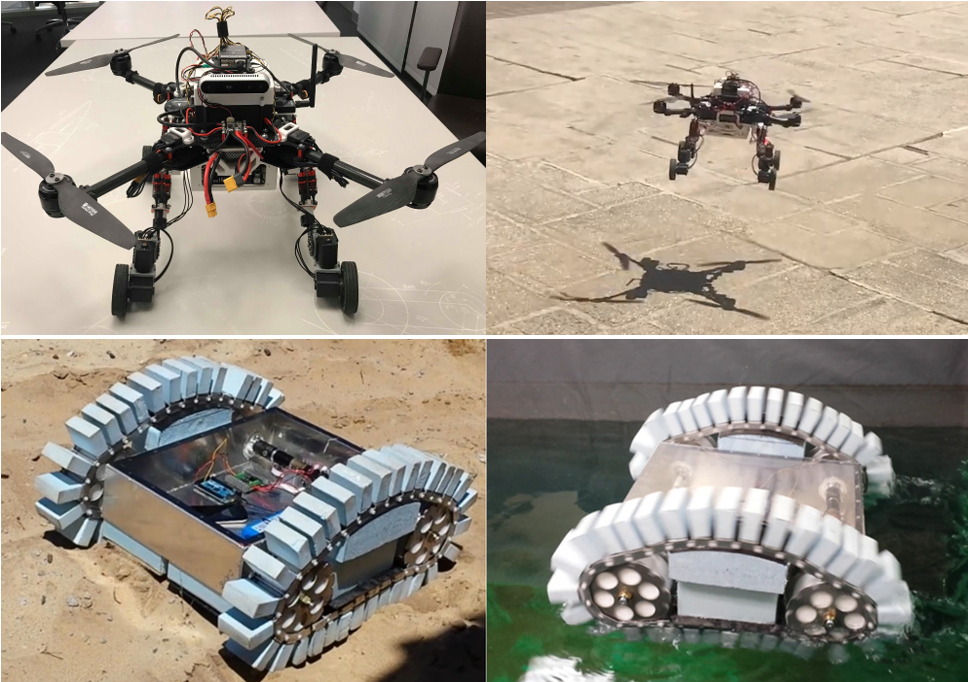}
    \vskip -0.1 true in
	\caption{Top Row: A "Drivocopter" Drone (developed by the authors) which can fly and drive. Video of operation can be accessed at \href{https://www.youtube.com/watch?v=QZyuvXfifvQ}{\textcolor{magenta}{https://www.youtube.com/watch?v=QZyuvXfifvQ}}. Bottom Row: "Ambot" Amphibious Robot \cite{Ambot} capable of ground and marine locomotion.}
	\label{fig:platformexamples}
\vskip -0.3 true in
\end{figure}


Other relevant work can be found in the concept of "Combined Task and Motion Planning" (TAMP) \cite{tamp1,tamp2,lgp}, in which a discrete sequence of tasks must be found simultaneously with the geometric motions that satisfy these sequences. Most of these works are in the domain of manipulation, where there are little-to-no continuous dynamic effects, which are often critical for locomotion. In addition, whereas each discrete task changes the geometric preconditions that affect the next task in a manipulation sequence, we assume a fixed geometric map in our framework. 

Within the field of robotic locomotion, there has been considerable effort to solve the "Hybrid Activity and Trajectory Planning" (HATP) problem \cite{hybridswarm,colin,scottactivity,IVS}, where individual tasks of multiple robots, as well as their trajectories, must simultaneously be planned.  Some of these planners also take into account the dynamic behavior of the robot. However, due to the difficulty of simultaneously doing trajectory generation with task planning, these approaches often consider simplified forms of dynamics, such as constant rate \cite{colin}, or first-order behavior \cite{scottactivity}, that admit fast and convex formulations of trajectory planning. By alleviating the need to compute exact control inputs in the discrete-planning process, and capturing their effects with offline-approximated costs, our approach allows tractable utilization of realistic robot dynamics. 

Our framework is most related to \cite{IVS}, which uses hierarchical planning for multi-agent systems: a graph-search first creates a global plan for multiple agents, and a local controller is used for the agents to track the global plan. The cost function in their graph utilizes the value function of the closed-loop policy that is computed offline, which is similar to our offline cost approximation. However, while \cite{IVS} utilizes distance-dependent energy costs with a double-integrator model for their robot dynamics, we directly optimize for electrical energy expenditure, thus capturing a more detailed and dynamically accurate behavior of the robot. This further emphasizes the full power of this framework.  

The proposed method is primarily implemented in simulation: the hybrid double-integrator with viscous friction is shown as a low-dimensional case (Sec.\ref{hybridd}). Then, example trajectories for more realistic systems are given by considering amphibious (Sec.\ref{amphibious}) and flying-driving locomotion (Sec.\ref{drivocopter}). As most hybrid planners (\cite{colin}, \cite{scottactivity}), our planner does not guarantee probabilistic optimality due to the underlying heuristics required to solve the problem, but we show that our method performs quantitatively well in practice.

\section{Problem Formulation}\label{framework}
\subsection{The Hybrid Locomotion System}
We define a hybrid locomotion system as a type of hybrid control system \cite{hybridcontrolsystem, xiaobin} with additional constraints. The hybrid locomotion system, $\mathscr{HL}$, is defined as a tuple  
\begin{equation}
    \mathscr{HL}=(FG, \mathcal{D},\mathcal{U},\mathcal{S},\Delta).
\end{equation}
\note{The "domain" definition was difficult to follow on the first reading, so I modified the description.  More importantly, the switching surface notation needs some improvement.  Switching surface mode $S_i$ is the boundary between what modes?  Should we have notation $S_{i,j}$ to denote the switching surface between $M_i$ and $M_j$? JWB}

In the following descriptions of each system element, $i$ indexes  the locomotion mode (i.e. flying or driving):
\begin{itemize}
    \item $FG=\{(f_i,g_i)\}$ describes the dynamics associated with each locomotion mode, which are assumed to take a control-affine form: $\dot{x}=f_i(x)+g_i(x)u$. 
    \item $\mathcal{D}=\{\mathcal{D}_i\}$ is the set of domains, or state-spaces, associated with the continuous dynamics of each mode.
    \item $\mathcal{U}=\{\mathcal{U}_i\}$ is the set of admissible control inputs associated with each mode.
    \item $\mathcal{S}=\{S_{i,j}\}$ is the set of guard surfaces that describes the boundaries between domains of mode $i$ and $j$.
    \item $\Delta=\{\Delta_{i\rightarrow j}\}$ is the set of reset maps that describe discrete transformations on the guard surface $S_{i,j}$  
\end{itemize}

We additionally assume that each state $x\in \bigcup \mathcal{D}_i$ belongs to a single mode $i$. i.e., the domains disjointly partition the reachable state-space. 

\subsection{Optimal Trajectories in the Hybrid Locomotion System}
To define an optimal hybrid trajectory, we formulate a cost for each mode's control-affine system in Bolza form:
\begin{equation}
    J_i=\Phi_i(x(t_0),t_0,x(t_f),t_f)+\int^{t_f}_{t_0}\mathcal{L}_i(x(t),u(t),t)dt.
\end{equation}
There also exists a constant switching cost $J(\Delta_{i\rightarrow j})$ to transition from one modality $i$ to $j$. We formulate the problem of finding the optimal trajectory for a hybrid locomotion system as the following two-point boundary value problem:
\begin{mini}|s|
{u}{\sum J_i+J(\Delta_{i\rightarrow j})}{}{}
\addConstraint{\dot{x}=f_i(x)+g_i(x)u \quad \forall x\in \mathcal{D}_i \quad u\in\mathcal{U}_i \quad \forall i}
\addConstraint{x(t_0)=x_0, \quad x(t_f)=x_f.}
\end{mini}
We want to find a trajectory that is dynamically feasible within each modality, while optimizing the cost functional throughout the entire trajectory, which would also require optimizing the order of discrete modes to visit. This problem can be transformed into an instance of a mixed-integer nonlinear program \cite{tobia}.

\section{Planning Methodology}\label{lowdimension}

Sec. \ref{sec:basic} summarizes our planning method that combines sampling-based planning with local trajectory optimization. To extend this approach to problems with high dimensions, Sec. \ref{sec:higher} introduces virtual constraints and cost approximation to improve real-time performance. 

\subsection{Dynamic Programming with Continuous Optimization} \label{sec:basic}

\subsubsection{Graph Structure} 
\begin{figure}[H]
\vskip -0.1 true in
	\centering\includegraphics[width = 0.45\textwidth]{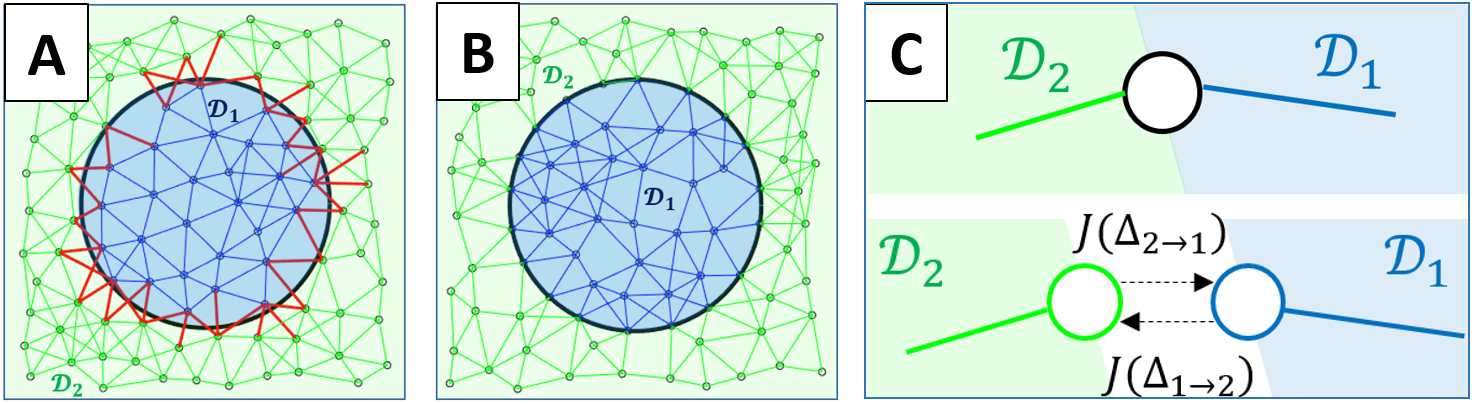}
	\vskip -0.1 true in
	\caption{A) red edges cross a guard surface between domains $\mathcal{D}_1$ and $\mathcal{D}_2$, violating the constraint on edges. B) By sampling on the guard surface and allowing no edges between $\mathcal{D}_1$ to $\mathcal{D}_2$, all graph edges are constrained to a single mode. C) Node augmentation to handle modality switching costs.}
	\label{fig:graphstructure}
\end{figure}

First, we discretize the problem by sampling coordinates in the free state-space of each domain $\mathcal{D}_i$. The vertices, $V$, of a digraph, $G(V,E)$, are constructed from these samples, similar to the framework of Probabilistic Roadmaps (PRM) \cite{prm}. The edges represent locally optimal paths between the vertices. Each edge is weighted with the optimal transport cost. To avoid the situation where the two vertices of an edge lie in different modalities, we additionally impose the following constraints:
\begin{enumerate}
    \item $e=(x_i\rightarrow x_j)\in E$, $x_i,x_j \in \mathcal{D}_k$ for some mode $k$. i.e., edges only connect states in the same mode. 
    \item We explicitly sample the guard surface and only allow paths to cross a guard through a guard sample point. 
\end{enumerate}
Fig.\ref{fig:graphstructure}. A and B illustrate these conditions. The shortest-path search is tackled by Djikstra's algorithm \cite{shortestpath} once the locally optimal trajectory costs are known. 
If there exists a switching cost from one modality to another, we augment the sample on the guard surface $x$ with two connected nodes $x_i$ and $x_j$ that shares the same state-space coordinates, and assign switching cost to the edge cost between the two samples, as illustrated in Fig. \ref{fig:graphstructure}.C.
\subsubsection{Continuous Optimization of Trajectory Segments}
As each edge connects states in a single mode, we assign the edge weight by solving the optimization problem:
\begin{mini}|s|
{u}{J_i}{}{w(x_1\rightarrow x_2)=}
\addConstraint{\dot{x}=f_i(x)+g_i(x)u}, \mkern4mu x\in\mathcal{D}_i, \mkern4mu u\in\mathcal{U}_i
\addConstraint{x(t_0)=x_1, \quad x(t_f)=x_2,}
\end{mini}
\note{The notation "$x_1,x_2\in M_i$" is poor.  It is not possible for a state to be in an index set.  INstead it {\em could} be "$x_1,x_2\in \mathcal{D}_i$}
where $x_1,x_2\in \mathcal{D}_i$. The result of this problem is used as the running cost in our dynamic programming framework. This standard trajectory optimization problem can be solved using existing methods, such as direct collocation \cite{gpops}.

\subsubsection{Final Path Smoothing}
The path(s) returned from graph-search are smoothed via trajectory optimization, knowing the switching sequence and the guard surface points. Given a path of samples $P=(x_1,x_2,\cdots, x_k)$ resulting from graph search, we partition the samples using their modalities: 
\begin{equation}
    \begin{split}
    \bigcup P_i & = 
        P_1 = \{x_i | 0 \leq i \leq k_1, \forall x_i \in \mathcal{D}_{j_1}\}\mkern9mu\cup\cdots\cup\\
        & P_n = \{x_i | k_{n-1} \leq i \leq k_n, \forall x_i \in \mathcal{D}_{j_n}\}, 
    \end{split}
\end{equation}
where $j_i$ denotes the mode of each partition, and $x_{k_i}$ denotes the sample on the guard surface where the trajectory switches modes. The optimal trajectories between boundary points are then found by re-solving the partition-wise trajectory optimization problem. The total trajectory is reconstructed by concatenating the partition-wise optimal trajectories. Existing works have shown that this type of final smoothing, which delays the final choice the control inputs until the end, often leads to much enhanced performance \cite{lgp,scottactivity}.


In addition, although we assume no significant presence obstacles in this work, we note that the result of the path of samples can be used as a nominal collision-free trajectory that can be used to bound the final smoothing process \cite{kumar1,kumar2}, which is a direction we further plan to investigate.

\subsection{Extension to High Dimensions}\label{sec:higher}
Although dynamic programming with running cost of continuous trajectory optimization shows good promise, it is computationally expensive, requiring $O(|V|^2)$ instances of trajectory optimization. In high dimensions, the number of samples increases exponentially if the resolution is maintained, and trajectory optimization methods scale poorly. The two methods introduced in this section aim to make this method tractable for high-dimensional systems.

\subsubsection{Virtual Constraints for Search-Space Reduction}
We can reduce the dimensionality of the sample space by introducing heuristic virtual constraints that fix some coordinates as functions of the sampled-coordinates. The state-space is divided into sampled coordinates ($x^s$) and auxiliary coordinates ($x^a$), which are functions of the sampled coordinates:
\begin{equation}
    x=(x^s,x^a)^T=\left(x^s,v\left(x^s\right)\right)^T.
\end{equation}
The state partioning into $x^s$ and $x^a$ is problem-dependent, but can be understood in the context of model-order reduction: if the original system and the virtually-constrained system show bounded difference in their evolution, it indicates a good choice of coordinates and constraints. Point-mass coordinates of position and velocity \cite{xiaobin}, or differentially flat coordinates \cite{minimumsnap} can be good choices. Eliminating the sampling of the subspace $x^a$ can significantly reduce computation, making the method tractable. 

\subsubsection{Approximate Dynamic Programming}
To find the weight between two sampled coordinates $x^s_1$ and $x^s_2$ in the graph, let us first define a function $J:\mathbb{R}^{\dim(x^s)}\times \mathbb{R}^{\dim(x^s)}\rightarrow \mathbb{R}$, which is described by the following optimization problem:
\begin{mini}|s|
{u}{J_i}{}{J(x_1^s,x_2^s)=}\label{opt:optim}
\addConstraint{\dot{x}=f_i(x)+g_i(x)u, \mkern9mu x\in\mathcal{D}_i, \mkern9mu u\in\mathcal{U}_i }
\addConstraint{x_0=[x_1^s, v(x_1^s)]^T, x_f=[x_2^s,v(x_2^s)]^T}
\end{mini}
where $(x_1^s, v(x_1^s))^T,(x_2^s, v(x_2^s))^T \in \mathcal{D}_i$. Since this optimization problem has to be solved $O(|V|^2)$ times, we choose to learn a function approximator offline.

Using $(x_1^s,x_2^s)$ as feature vectors, and $J(x_1^s,x_2^s)$ as label, we first produce a batch $\left(\left(x_1^s,x_2^s\right),J\left(x_1^s,x_2^s\right)\right)$ from multiple trajectory optimization runs. Then, function approximators from supervised learning algorithms such as Support Vector Regression (SVR) \cite{svr} or Neural Nets are used to approximate $J(x_1^s,x_2^s)$. Denoting the approximated function as $\tilde{J}(x_1^s,x_2^s)$, the weights on the graph are assigned by $w(x_1^s\rightarrow x_2^s)=\tilde{J}(x_1^s,x_2^s)$.

Since $\tilde{J}$ is learned offline, its evaluation does not require a full instance of nonlinear programming, greatly reducing online computation. Yet, as $\tilde{J}$ is learned from trajectory optimization, all costs in Bolza form can be utilized, and dynamic  or temporal constraints can be incorporated.

\section{Case Study: Hybrid Double Integrator}\label{hybridd}
This section first verifies our low-dimensional method for 1D problem of a thrust-vectored mass on a linear rail, with viscous drag appearing at $p\geq 0$. This can be formulated as a hybrid locomotion system with the dynamics of: 
\begin{equation}
    \ddot{p}=u \text{ if } p < 0, \quad \ddot{p}=u - \dot{p} \text{ if } p \geq 0.
\end{equation} 
In addition, consider that we have the input constraint $|u|\leq 1$ for both domains. Converting this to a first-order system $x=\begin{pmatrix} p, v \end{pmatrix}^T$, the system can be described as:
\begin{equation}
    \mathscr{HL}=\begin{cases} 
        FG & =\{(f_-,g_-),(f_+,g_+)\},\\\mathcal{D}&=\left\{\left\{x|p<0\right\},\left\{x|p\geq 0\right\}\right\} \\ \mathcal{U}&=\left\{\left\{u||u|\leq 1\right\},\left\{u||u|\leq 1\right\}\right\} \\ \mathcal{S}&=\left\{S_{+,-}=\left\{x|p=0\right\}\right\},\\ 
        \Delta& = \left\{\Delta_{+,-}= x^+\rightarrow x^-\right\}
    \end{cases},
\end{equation}
where the dynamics are described by,
\begin{equation}
    f_- = \begin{pmatrix} 0 & 1 \\ 0 & 0 \end{pmatrix} \; , \;
    f_+ = \begin{pmatrix} 0 & 1 \\ 0 & -1 \end{pmatrix} \; , \; g_+ = g_-=\begin{pmatrix} 0 \\ 1 \end{pmatrix}. 
\end{equation}
Then, let us find a trajectory from $x_i$ to $x_f$ while minimizing the input, 
\begin{equation}
    J_- = J_+ = \int^{t_f}_{t_0} u^2 dt. 
\end{equation}

\begin{figure}[H]
	\centering\includegraphics[width = 0.5\textwidth]{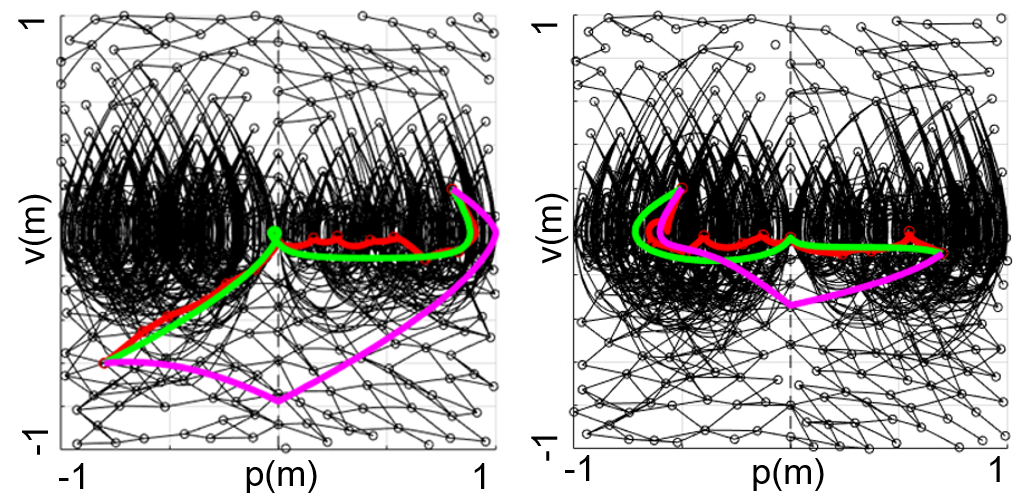}
	\caption{Optimal Trajectories from $x_i=[0.8,0.2]^T$ to $x_f=[-0.8,-0.]^T$ (left), and from $x_i=[0.7,-0,1]^T$ to $x_f=[-0.5,0.2]^T$ (right). Red trajectories are obtained using graph search, green trajectories are results of final smoothened path, and the pink trajectory is result of Multi-phase optimization in GPOPS-II \cite{gpops}. The edges represent optimal trajectories between each sample.}
	\label{fig:hybridintegrator}
\end{figure}

Using our framework, we first place a graph structure on the state-space using knowledge of the domains $\mathcal{D}_i$, then optimize each continuous trajectory using GPOPS-II \cite{gpops} with IPOPT \cite{ipopt} solver. The trajectory obtained using graph search, and the final smoothened trajectory using the knowledge of the switching sequence and the boundary points on the guard surface is displayed in Fig.\ref{fig:hybridintegrator}. 

Finally, since the switching sequence is trivial to guess for this example, we utilize multi-phase optimization in GPOPS-II with IPOPT, which puts an equality constraint from the end of the first phase in $\mathcal{D}_+$ and the beginning of the second phase in $\mathcal{D}_-$ and compare the results. The trajectory using multi-phase optimization is displayed in Fig.\ref{fig:hybridintegrator}.

\begin{figure}[H]
	\centering\includegraphics[width = 0.5\textwidth]{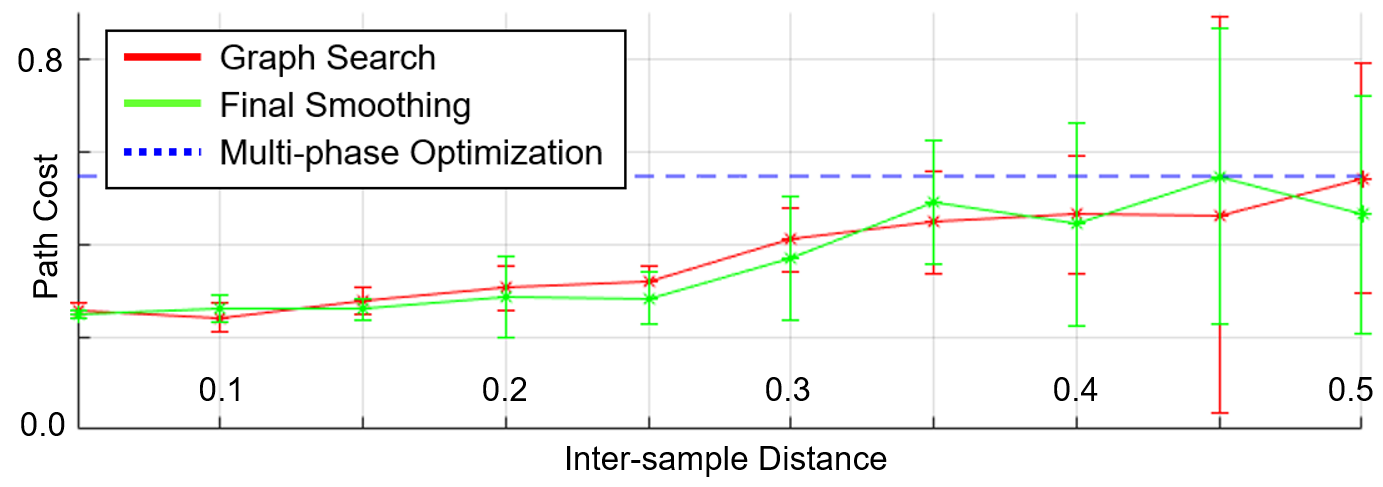}
	\caption{Empirical convergence of cost with decreasing intersample distance.}
	\label{fig:hybridintegratordata}
\end{figure}

To empirically study the effect of having increased number of samples, we run the algorithm $10$ times with different inter-sample distances (controlled by Poisson disc sampling \cite{poissonsampling} on the state-space), and show convergence in Fig. \ref{fig:hybridintegratordata}. Fig. \ref{fig:hybridintegratordata} shows that our method results in a lower cost compared to multi-phase optimization, with inter-sample distance as large as $0.3$. Although the dynamics are linear and cost is quadratic, the problem is no longer convex in the switching coordinate. Thus, our PRM framework, which searches more globally over the domain, performs better than the local optimum provided by IPOPT \cite{ipopt}.

\section{2D Case Study: Amphibious Tank (AmBot)} \label{amphibious}
This section describes optimal trajectories for the amphibious vehicle introduced in \cite{Ambot}, which uses tank treads for ground locomotion (skid-steer), and marine locomotion (paddles). After describing the vehicle dynamics in both modes, we obtain optimal trajectories for an example environment. 

\subsection{Dynamics}
\subsubsection{Ground and Marine Dynamics}
We derive the Newtonian mechanics for planar motion, and incorporate first-order armature motor dynamics. The ground states, $x_g\in\mathbb{R}^{6}$, and marine states, $x_m\in\mathbb{R}^8$, are defined as
\begin{equation}
    \begin{cases}
        x_g&=(p^w_b, v^b,\theta^w_b,\omega^b)^T\\
        x_m&=(p^w_b,v^b, \theta^w_b, \omega^b,\phi_L,\phi_R)^T,
    \end{cases}
\end{equation}
where $p^w_b\in \mathbb{R}^2$ is the body position with respect to (wrt) a world frame, $v^b\in\mathbb{R}^2$ is the velocity in the body frame, and $\theta^w_b,\omega^b\in \mathbb{R}$ denote the orientation and angular velocity wrt a world frame. Finally, $\phi_L,\phi_R\in\mathbb{R}$ denote the left and right motor speeds. In both locomotion modes, the control action $u_g=u_m=(u_L,u_R)^T\in [-1,1]^2$ correspond to commanded motor speeds via fraction of applied motor voltage. 

We model a no-slip constraint for ground operation. A $1^{st}$-order motor model relates motor torque (which generates tractive forces on the vehicle) to command inputs. A drag force proportional to the square of vehicle speed and a similar $1^{st}$-order motor model are used in the aquatic domain. 

\subsubsection{Hybrid Dynamics}
The governing dynamical systems for each mode are represented by the hybrid dynamics 
\begin{equation}
    FG=\begin{cases}
        \dot{x}_g = f_g(x_g)+g_g(x_g)u_g & x \in \mathcal{D}_g \\ 
        \dot{x}_m = f_m(x_m)+g_m(x_m)u_m & x \in \mathcal{D}_m, 
    \end{cases}
\end{equation}
where $g,m$ denotes ground and marine modes. The domains and guard surfaces $\mathcal{D}_{g},\mathcal{D}_m,\mathcal{S}_{m,g}$ are obtained from terrain, and $\mathcal{U}_g=\mathcal{U}_m=[-1,1]^2$ for both inputs. We apply the identity map to  $\Delta_{m\rightarrow g}$ and $\Delta_{g\rightarrow m}$.

\subsubsection{Cost Function}
We minimize the robot's total energy expenditure, modeled as:
\begin{equation}    \label{eq:groundcost}
    J_g=J_m= \int^{t_f}_{t_i} \bigg(\sum_{i=L,R}V_{cc}u_i\cdot \frac{k_t}{R}(V_{cc}u_i-k_t\phi_i)+P_d \bigg)dt,
\end{equation}
where $V_{cc}$ is the battery voltage, $k_t$ is the motor torque-constant, $R$ the internal resistance, and $P_d$ the constant power drain. The first term models actuator power dissipation, and the latter term models constant power drainage from on-board electronics. We assume no switching costs associated with the discrete reset map, $J(\Delta_{m\rightarrow g})=J(\Delta_{g\rightarrow m})=0$ 
\subsection{Learning the Cost Approximator}
For ground operation, we divide the state-space $x_g\in\mathbb{R}^6$ into sampled and auxiliary coordinates 
\begin{equation} \label{eq:groundstates}
    x^s_g = (p^w_x,p^w_y,v^b_x,\theta)^T ,\, x^a_g =(v^b_y,\omega)^T = (0,0)^T.
\end{equation} 
This heuristic division of coordinates recognizes that side-slip is constrained for skid-steer vehicles, and angular velocity is small. 
\begin{figure}[t]
	\centering\includegraphics[width = 0.45\textwidth]{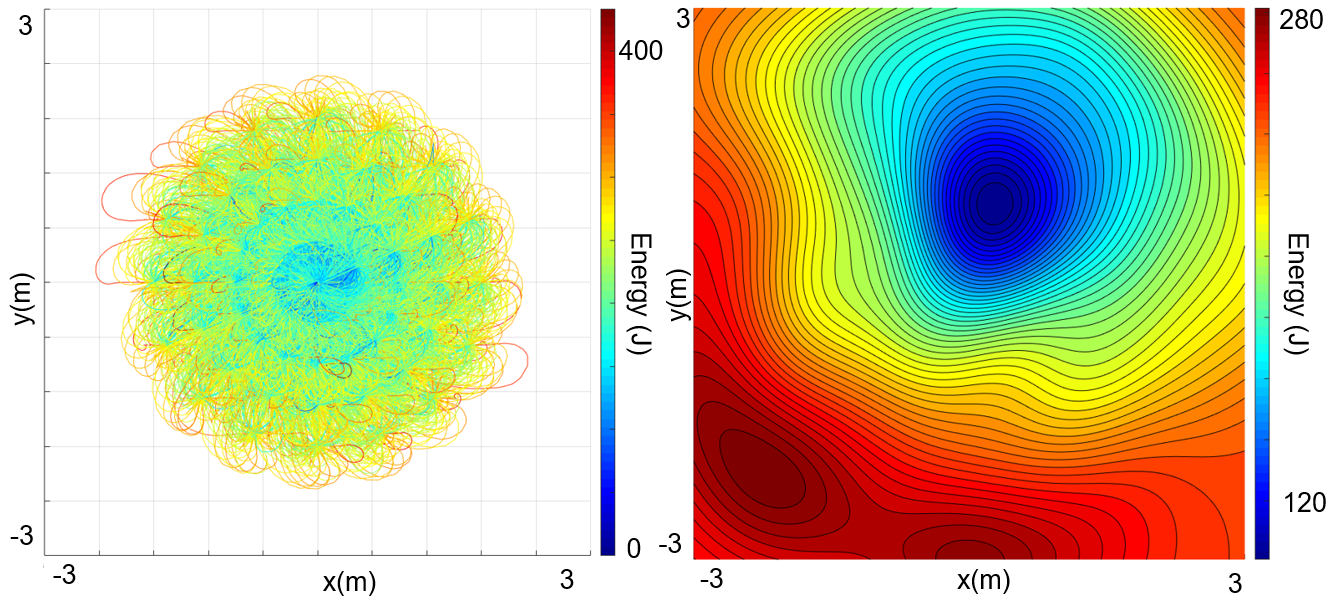}
	\vskip -0.1 true in
	\caption{Left: 11520 ground trajectories colored by their cost. Right: $xy$-energy contour for $v^b_x=1.0m/s$, $\theta=\pi/2$. The heatmap corresponds to the energy cost to go from $x_i^s=[0,0,0,0]^T$ to $x_f^s=[x,y,1.0,\pi/2]^T$. The approximated cost captures complex  effects of underactuation.}
	\label{fig:trajectorylearning}
		\vskip -0.25 true in
\end{figure}
For marine operation, $x_m$ is divided into 
\begin{equation}
    \begin{cases}
    x^s_m = (p^w_x, p^w_y, v^b_x, \theta)^T \\ 
    x^a_m = (v^b_y, \omega, \phi_L, \phi_R)^T = (0, 0, \phi_n,\phi_n)^T,
    \end{cases}
\end{equation}
where track forces equal water drag at equilibrium speed $\phi_n$. 

\note{The translation invariance in the next paragraph is hard to understand. JWB}

The cost $J(x_i^s,x_f^s)$ in \eqref{opt:optim} is approximated from multiple optimal trajectories evaluated offline. Using $11520$ samples, the function $J(x_i^s,x_f^s)$ is evaluated using GPOPS-II \cite{gpops}, and SVR with Gaussian kernel trains the function $\tilde{J}$ with Sequential Minimal Optimization \cite{SVM}. The process is repeated for ground and marine locomotion. Fig.\ref{fig:trajectorylearning} shows the resulting trajectories and the learned function's contour. 


\subsection{Results}
\begin{figure}[H]
\vskip -0.1 true in
	\centering\includegraphics[width = 0.47\textwidth]{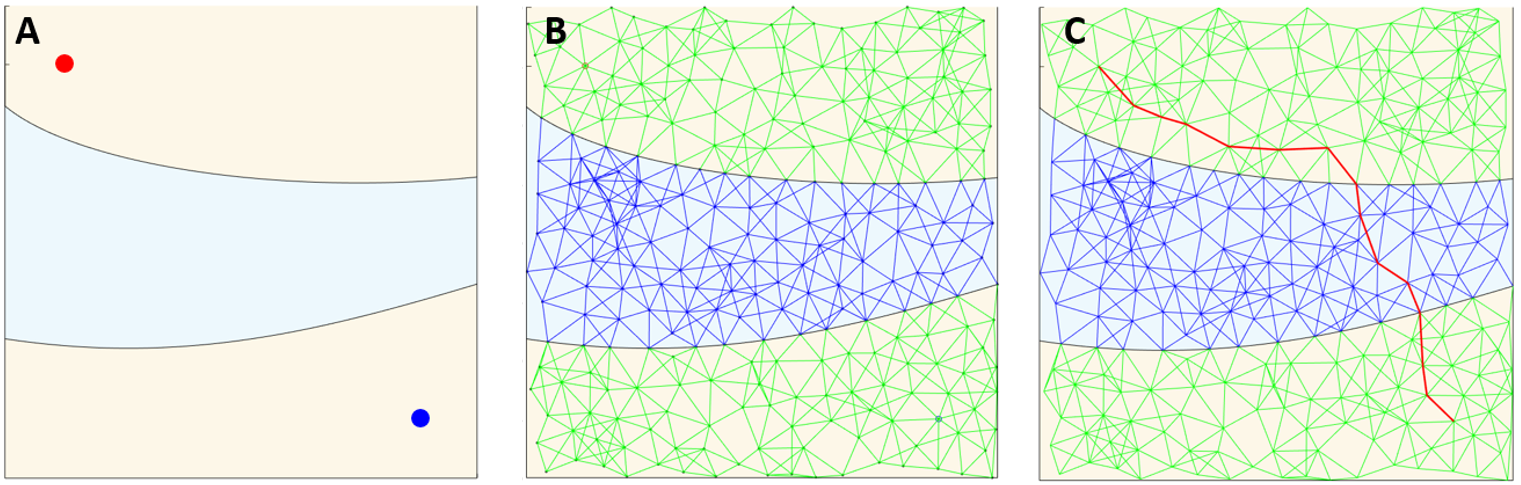}
	\vskip -0.15 true in
	\caption{A: Model Environment. B: Graph Generation. C. Result of shortest path search.}
	\label{fig:amphibiousframework}
\end{figure}

\begin{figure}[H]
\vskip -0.2 true in
	\centering\includegraphics[width = 0.47\textwidth]{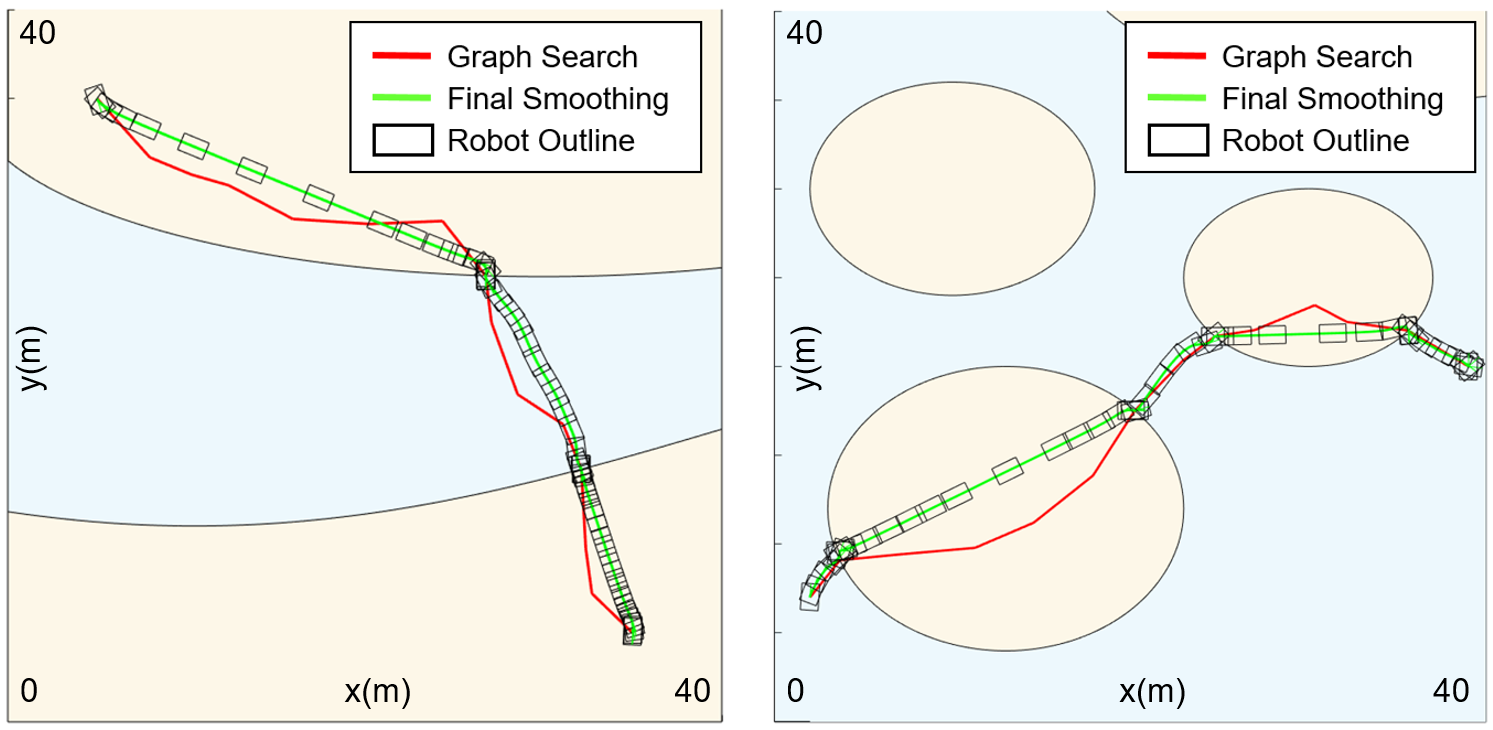}
	\vskip -0.1 true in
	\caption{Final trajectories for example of river crossing (left), and island crossing (right). The Robot outline is displayed at equal time differences.}
	\label{fig:amphibiousfinal}
\end{figure}

We sample position using the method of Sec.\ref{lowdimension}.A, and grid the states $v_x,\theta$ to create $x^s$. The edge weights are estimated from the learned function $\tilde{J}$. Finally, the shortest path is found by Djikstra's algorithm \cite{shortestpath}. Fig. \ref{fig:amphibiousframework} illustrates this process. The final smoothed trajectory is shown in Fig.\ref{fig:amphibiousfinal}. 

The final trajectories differ noticeably from those produced by a shortest-path planner due to the differences in Costs of Transport. Since the robot expends more energy in water, it drives further on the ground until it switches to swimming. This example shows that our method exhibits reasonably correct qualitative behavior.

%

\begin{table*}[!b]
\vskip -0.1 true in
  \centering
  \begin{tabular}{|r|rrrrrr|rr|}
    \hline
     & \multicolumn{6}{c|}{Cost from Fixed Sequence Heuristic Trajectories (Joules)} & \multicolumn{2}{c|}{Our Method} \\\hline 
    Dist. (m) & F & DF & DFD & FDF & DFDF & DFDFD & Sequence & Cost (Joules)\\\hline 
    110 & 12129.72 & 11991.18 & 11780.26 & \textbf{6612.62} & 6620.16 & 6642.41 & DFDFD & 6655.44\\ 
    90 & 10167.32 & 9902.18 & 9867.10 & \textbf{6345.64} & 6558.94 & 6578.74 & DFDFD & 6600.45\\ 
    70 & 8208.72 & 7941.78 & 7726.73 & \textbf{6470.98} & 6492.85 & 6511.88 & DFD & 8156.85\\ 
    50 & 6248.72 & 6112.78 & \textbf{5894.58} & 6477.60 & 6449.91 & 6452.32 & DF & 6456.37\\ 
    30 & 4244.2 & 4148.18 & \textbf{3815.68} & 6343.64 & 6365.65 & 6387.47 & DFD & 4033.92\\\hline 
  \end{tabular}
  \caption{Comparison of costs from heuristic trajectories with fixed sequences. D: Driving, F: Flying}
  \label{tab:1}
\vskip -0.125 true in 
\end{table*}
\begin{figure*}[!b]
\vskip -0.125 true in
	\centering\includegraphics[width = 1.0\textwidth]{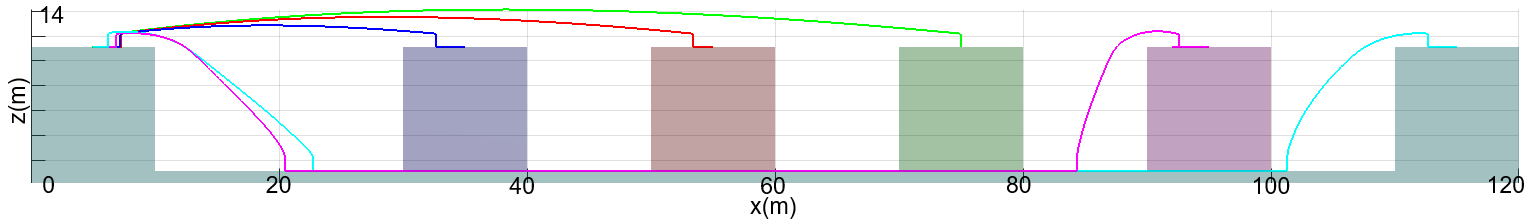}
	\vskip -0.1 true in
    \caption{Depiction of trajectory differences as the target platforms become more distant. At less than $75m$ separation, the robot always flies. After $95$ meters separation,  driving in the chasm saves energy.}
	\label{fig:4traj}
\end{figure*}

\section{3D Case Study: Drivocopter} \label{drivocopter}

This section models the {\em Drivocopter} flying-driving drone of Fig. \ref{fig:platformexamples}.  It uses skid-steer driving and quadrotor flight. 

\subsection{Dynamics}
We use the ground model of Sec.\ref{amphibious} with different parameters, while the flight dynamics are based on \cite{omnicopter} and \cite{minimumenergy}.
\subsubsection{Flight Dynamics}
Standard rigid-body dynamics \cite{quadrotordynamics} describe flight motions driven by four rotor forces, which use a speed-squared-dependent lift term and $1^{st}$-order armature motor dynamics. The state vector $x_f\in\mathbb{R}^{16}$ is 
\begin{equation}
    x_f= (p^w_b, v^b, \Theta^w_b,\omega^b,\phi_i)^T, 
\end{equation}
where $p^w_b\in\mathbb{R}^3$ is the vehicle position wrt a world frame, $v^b\in\mathbb{R}^3$ is the 3D velocity in the body frame, $\Theta^w_b\in\mathbb{R}^3$ denotes vehicle orientation w.r.t the world frame, parametrized by ZYX Euler angles, $\omega^b\in\mathbb{R}^3$ is the body angular velocity, and $\phi_i=(\phi_1,\phi_2,\phi_3,\phi_4)\in\mathbb{R}^4$ are the motor rotational speeds. 

\subsubsection{Hybrid Dynamics}
Again, the two modalities of ground and flight are represented by a hybrid dynamical system 
\begin{equation}
    FG=\begin{cases}
        \dot{x}_f = f_f(x_f)+g_f(x_f)u_f & x_f\in \mathcal{D}_f \\ 
        \dot{x}_g = f_g(x_g)+g_g(x_g)u_g & x_g \in \mathcal{D}_g ,
    \end{cases}
\end{equation}
where $f,g$ denotes flight and ground modes, the domains and guard surfaces $\mathcal{D}_{f},\mathcal{D}_g,\mathcal{S}_{f,g}$ are obtained from the ground surface geometry. The motor inputs are $\mathcal{U}_f=(u_1,u_2,u_3,u_4)=[0,1]^4$ with $\mathcal{U}_g=(u_L,u_R)=[-1,1]^2$. Finally, $\Delta_{f\rightarrow g}$ (landing) and $\Delta_{g\rightarrow f}$ (takeoff) are discrete transitions:
\begin{equation}
    \begin{cases}
        \Delta_{f\rightarrow g}=(p^w_x, p^w_y, p^w_z, 0^{9}, \phi_n)\rightarrow (p^w_x, p^w_y, 0^{4}) \\ 
        \Delta_{g\rightarrow f}=(p^w_x, p^w_y, 0^{2}, \theta^w_b, 0)\rightarrow (p^w_x, p^w_y, p^w_z, 0^{9}, \phi_n),
    \end{cases}
\end{equation}
where $\phi_n$ is the motor speed needed to provide hovering lift. During takeoff, we set $p^w_z$ to be a meter higher than the ground surface of the ground sample.
\subsubsection{Cost Function}
We use the same ground energy cost as \eqref{eq:groundcost}, and formulate the same energy for flight with different motor parameters. The costs for reset maps $J(\Delta_{f\rightarrow g})$ and $J(\Delta_{g\rightarrow f})$ are constant takeoff and landing energy costs obtained via trajectory optimization. 

\subsection{Learning the Cost Approximator}
\begin{figure}[H]
\vskip -0.15 true in
	\centering\includegraphics[width = 0.47\textwidth]{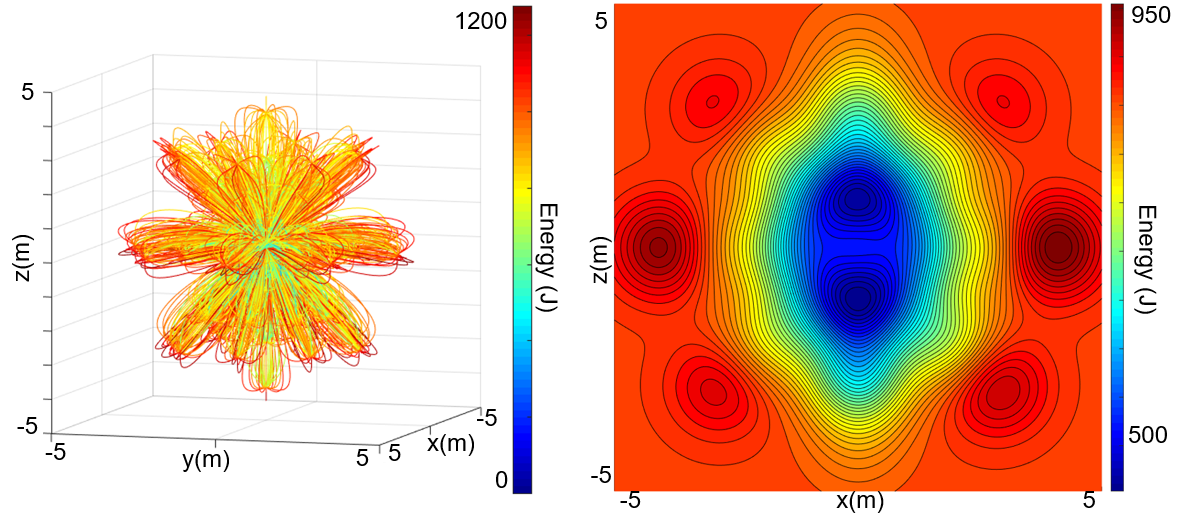}
	\vskip -0.1 true in
	\caption{Left: $17016$ trajectories produced to learn the flight energy function. Right: xz-projection of the learned function $\tilde{J}(0,x)$. 
	}
	\label{fig:resultflight}
\end{figure}
The ground states are divided into sampled / auxiliary coordinates via \eqref{eq:groundstates}. Flight states are divided by: 
\begin{equation}
    \begin{cases}
    x^s_f=(p^w_b, v^b)^T \\ x^a_f=(\Theta^w_b,\omega^b,\phi_i)=(0^{1\times 3},0^{1\times 3}, \phi_n \cdot 1^{1\times 4})^T,
    \end{cases}
\end{equation}
where the $\phi_n$ is the rotor rate at which the lift provided by the propellers allows the drone to hover in stable equilibrium. The cost $J(x_1^s,x_2^s)$ is learned as in Sec.\ref{amphibious}.B from $17016$ paths. Fig. \ref{fig:resultflight} shows the trajectories and energy map. The ground energy cost is found with Drivocopter parameters. 

\subsection{Results}
\begin{figure}[H]
\vskip -0.1 true in
	\centering\includegraphics[width = 0.45\textwidth]{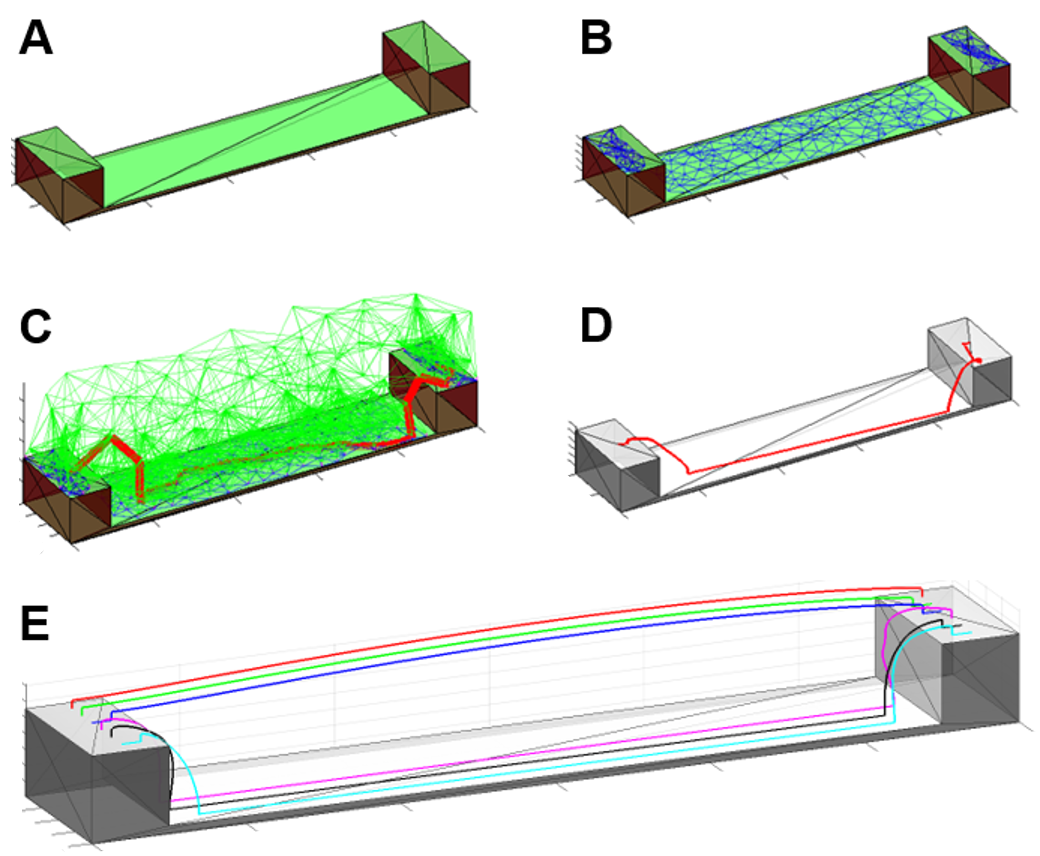}
	\vskip -0.15 true in
	\caption{A. Model Terrain classified into drivable and undrivable terrains. B. Poisson sampling on ground mesh. C. Poisson sampling on air and shortest path search. D. Smoothened final path. E. Heuristic trajectories for comparison in Tab.\ref{tab:1}. From back to front: F (red), DF (green), DFD (blue), FDF (pink), DFDF (black), DFDFD (cyan)}
	\label{fig:drivofinal}
	\vskip -0.1 true in
\end{figure}

A CAD environment model, consisting of two raised platforms separated by a flat-bottom chasm, is meshed into drivable and undrivable regions (Fig.\ref{fig:drivofinal}.A), and the ground and free-space meshes are Poisson sampled (Fig.\ref{fig:drivofinal}.B). The result of a shortest-path (Fig.\ref{fig:drivofinal}.C) is smoothened (Fig.\ref{fig:drivofinal}.D). This process is depicted in Fig. \ref{fig:drivofinal}. We hypothesized that when the landing platforms are nearby, the drone should not drive in the chasm, since gravitational losses exceed energy gains by driving. As the landing platform becomes more distant, the drone saves energy by driving in the chasm.  We tested this idea on 5 different terrains parametrized by the distance between platforms (see Fig.\ref{fig:4traj}). Our planning results show correct qualitative behavior. Illustration of the trajectories is available in the video \cite{video}.

To illustrate how our results perform quantitatively, we also generate a few heuristic trajectories with different sequences (illustrated in Fig.\ref{fig:drivofinal}.E) and compare the final cost of these heuristic trajectories with our method in Table.\ref{tab:1}. Our method produces switching sequences that mostly agree with lowest-cost producing sequences among heuristic trajectories, and costs are quantitatively comparable to the heuristically optimal trajectories. However, our comparison is limited by the fact that the true optimal solution to the original mixed-integer problem is not tractable to obtain.

\section{Conclusion} \label{conclusion}
We presented a novel scheme to plan energy-efficient hybrid locomotion trajectories using approximate dynamic programming. Through capturing optimal policies within individual modalities with the optimal cost function, we showed that our approach is successful in decoupling the continuous and discrete optimization problems. We have also demonstrated that our approximated cost is successful in capturing complex dynamic characteristics of the robot through examples of practical hybrid locomotion: the hybrid double-integrator, the Ambot, and the Drivocopter. 

Improvements are possible by upgrading elements of this framework. Better computational speed could be realized by adaptive sampling \cite{hsu}. An (A*) \cite{astar} graph search would be enabled by transport energy heuristics, while other function approximators, such as Neural Nets, might improve the cost function learning module. More efficient implementations \cite{FORCESNLP} can be used for trajectory optimization.

We also note some major limitations of the planner. Our framework relies on approximating solutions to Boundary Value Problems (BVP), but it is difficult to guarantee how well the function approximator captures the cost landscape, especially due to reachability constraints; even if the resulting BVP is infeasible, the cost approximator will still return a finite cost. An RRT \cite{rrt} approach to planning through hybrid dynamical systems can address this issue \cite{R3T}; while all nodes are reachable in this approach, detecting where the tree has crossed the guard surface is much more difficult.

For future works, we wish to understand how this planner can extend to settings with many obstacles, by using our sampled points as waypoints that can bound the result of final path smoothing. In addition, we are interested in studying cases where the map is not known in advance \cite{temporal}. Finally, efforts are underway to demonstrate our results experimentally on the Drivocopter of Fig. \ref{fig:platformexamples}. 

\bibliographystyle{IEEEtran}
\addtolength{\textheight}{-5.0cm}
\bibliography{thesisbib2}
\end{document}